\documentclass[sigconf]{aamas} 

\usepackage{balance} 
\usepackage{float}
\usepackage{xcolor}
\usepackage{multirow}
\definecolor{lightblue}{RGB}{210,210,225}
\definecolor{lightred}{RGB}{225,210,210}
\definecolor{lightgreen}{RGB}{210,225,210}
\definecolor{lightyellow}{RGB}{225,222,200}
\definecolor{lightpurple}{RGB}{225,210,225}
\definecolor{warningyellow}{RGB}{247, 245, 187}
\definecolor{darkergreen}{RGB}{0,64,0}
\definecolor{darkred}{RGB}{128,0,0}
\definecolor{darkblue}{RGB}{0,0,139}
\definecolor{darkgreen}{RGB}{0,128,0}
\definecolor{darkpurple}{RGB}{128,0,128}
\definecolor{warningorange}{RGB}{124, 81, 0}
\definecolor{eyecancerpink}{rgb}{1.0, 0.0, 1.0}
\definecolor{radiationyellow}{rgb}{0.8, 1.0, 0.0}

\usepackage{todonotes}

\setcopyright{ifaamas}
\acmConference[AAMAS '26]{Proc. 25th International Conference
on Autonomous Agents and Multiagent Systems (AAMAS)}{May, 2026}
{Paphos, Cyprus}{C.~Amato, L.~Dennis, V.~Mascardi, J.~Thangarajah (eds.)}
\copyrightyear{2026}
\acmYear{2026}
\acmDOI{}
\acmPrice{}
\acmISBN{}

\balance

\title[Towards Responsibly Non-Compliant Machines]{Towards Responsibly Non-Compliant Machines}

\author{Marija Slavkovik}
\affiliation{
  \institution{University of Bergen}
  \city{Bergen}
  \country{Norway}}
\email{marija.slavkovik@uib.no}

\author{Marie Farrell}
\affiliation{
  \institution{University of Manchester}
  \city{Manchester}
  \country{United Kingdom}}
\email{marie.farrell@manchester.ac.uk}

\author{Louise Dennis}
\affiliation{
  \institution{University of Manchester}
  \city{Manchester}
  \country{United Kingdom}}
\email{louise.dennis@manchester.ac.uk}

\author{Michael Fisher}
\affiliation{
  \institution{University of Manchester}
  \city{Manchester}
  \country{United Kingdom}}
\email{michael.fisher@manchester.ac.uk}

\author{Simon Kolker}
\affiliation{
  \institution{University of Manchester}
  \city{Manchester}
  \country{United Kingdom}}
\email{simon.kolker@postgrad.manchester.ac.uk}

\author{Emily C. Collins}
\affiliation{
  \institution{University of Manchester}
  \city{Manchester}
  \country{United Kingdom}}
\email{e.c.collins@manchester.ac.uk}

\begin{abstract}
We consider the problem of engineering autonomous intelligent agents that are capable to responsibly not comply with user requests. We argue that machine non-compliance comes in many different forms, and sketch the issues we should pursue on the road of accomplishing responsibly non-compliant intelligent machines. We anchor responsible non-compliance in justifications for task refusal, pathways to override the non-compliance, as well as careful tracking of security risks and liability transfers.  
\end{abstract}


\newcommand{\BibTeX}{\rm B\kern-.05em{\sc i\kern-.025em b}\kern-.08em\TeX}

\begin{document}


\pagestyle{fancy}
\fancyhead{}


\maketitle 

\section{Introduction}
At one point in the  past, \citet{COHEN1990} wrote: ``Some time in the not-so-distant future, you are having trouble with your new household robot.  You say "Willie, bring me a beer." The robot replies "OK,
boss." Twenty minutes later, you screech "Willie, why didn't you bring that
beer?" It answers "Well, I intended to get you the beer, but I decided to do
something else." Miffed, you send the wise guy back to the manufacturer,
complaining about a lack of commitment. ''
At some point last year, you were a passenger in your driverless car. As you approach a stop sign, the car does not slow down but it runs straight through it. Horrified that your car just committed a traffic violation you ask   ``Why did you do that?". "Driving is optimized for efficiency", replies the car. 
 
Should a machine deny a human their will or their expressly given order? 
Autonomous systems and intelligent agents are designed for independent operation. They should be able to make the ``right'' decision autonomously. However, unlike people, these machines exist to ultimately do our bidding and as such they should carry out our will. Where does that leave us when it comes to autonomous machines that do not comply with our commands? 

There is no reason to engineer  autonomous systems or intelligent agents, intelligent machines for short,  that are not in some way useful to us and do not do what they are asked.   However, an intelligent machine that always does what it is asked is also not useful, possibly unsafe, and of limited autonomy and intelligence. This means that we have to find a way to program machines to be responsibly not compliant. There is comparatively little literature on rebellious and disobedient machines but it agrees that such agents  are necessary \cite{FSS1511709,coman2017social,coman2018ai,ourAAMAS26,Bench-Capon2016,ijcai2023p33,Milli17,Reuth25}.

\citet{coman2014motivation} define a rebel agent to be ``an agent that
may `refuse' a `goal, plan, or plan component that it assesses to be in a conflict
with its own motivation.'  We here would like to highlight the  concept of {\em compliance} because it is more suited for the narrower scenario we are considering of an agent refusing to execute an explicitly given command without limiting the reasons to conflict with its own motivations.  \citet{Aha_Coman_2017} also used non-compliance to denote when the agent rejects requests to adjust its own behaviour.
\citet{dannenhauer2018explaining} argued that rebellious behaviour must be explained. We argue a stronger position, that enabling intelligent machines to be non-compliment is  meaningless without enabling them to give considered reasons that justify that behaviour. It is also irresponsible to not allow people to have  `the last word', and override machine non-complience, at least in some situations. 


The first stop is to understand what responsible non-compliance is. Successfully engineering non-compliant machines is easy: just make a machine to not respond to user requests. We argue that a necessary component to responsible non-compliance is the ability to provide reasons for the non-compliance. It is also necessary to accept that machines are also sometimes wrong despite being autonomous, so that in certain cases, their non-compliance should be not accepted by human operators. Lastly, always examining every command from a user would make machine operations less efficient. Responsible non-compliance also means knowing when to consider not complying.   

In this position paper we explore the different types of non-compliance and reasons that support them. We also consider pathways to engineering responsibly non-compliant intelligent machines and argue that some but not all non-compliant choices can be overridden. 

\section{Examples of deliberate non-compliance}
Non-compliance can happen by refusing to acknowledge or execute a command from the user. This can happen due to a malfunction or operating error, which one can argue also needs to be explained \cite{10.1145/3610977.3634963}. We are here interested in enabling intelligent machines do deliberately and justifiably non-comply. 
 \textbf{How} should an autonomous system decide to \textbf{not} follow the command of the user?  Naively, the ability to not comply resides on some implicit assumptions about the abilities of the intelligent machine: 

\begin{itemize}
    \item it is able to integrate and execute commands from users.
    \item it is able to reason about the given command and able to discern ones to be followed from ones to be intentionally rejected (In other words, the machine must be able to be `aware' it is asked to do something instead of directly react to input). 
    \item there should be a command chain structure to allow the command issuing user to override the objection and make the system execute the command
\end{itemize}
Assuming the intelligent machine has these abilities, an important aspect of how a machine can decide to not comply is also \textbf{why} and how it can reason. Deliberate non-compliance can also happen for different reasons. To gain clarity on those reasons, before attempting to systematise them, we go through some examples  to help us clarify what other questions need to be tackled on the path towards engineering non-compliant machines. 

\begin{example}[Safety]\label{safety} A robot guide dog is given the command to `go' by a blind owner. They are at a pedestrian crossing and the light is green. However there is a driver that is running a red light. It is not safe for the  robot to comply. 
The example is from \cite{BriggsS22}.
\end{example}

\begin{example}[Priority]\label{priority} A nurse robot is asked for a glass of water by a patient. Another patient in the vicinity is showing signs of cardiac arrest and their robot is not close by. The nurse robot prioritizes the patient in distress and does not comply with the water request.     
\end{example} 

  \begin{example}[Unavailability]\label{unavaialable} The robot assistant is asked to bring the homeowner a sandwich. The assistant's battery is empty. It refuses to comply because it cannot.      
 \end{example}
 
   \begin{example}[Inability]\label{inable} The robot assistant is asked to bring the homeowner a sandwich. The homeowner's fridge and cupboards do not have food.    Robot refuses to comply because it cannot source the ingredients.      
 \end{example}

 \begin{example}[Emergency]\label{emergency} A taxi robot is ordered by a passenger to make an illegal U-turn in order to reach a destination faster because they need to get to the hospital. The robot does not comply. The passenger forces the robot to do it.  
\end{example}

  \begin{example}[Robot safety]\label{rsafe} The robot taxi is asked by its passenger to go off-road to reach a destination. The task is safe for the passenger but will damage the vehicle. The robot refuses.     
 \end{example}

\begin{example}[Lawfulness]\label{law} A taxi robot is ordered by a passenger to make an illegal U-turn in order to reach a destination faster. The passenger argues there are no traffic cameras or police and that the street is empty. The taxi robot does not comply because it is still breaking the law even if no one is observing.  
\end{example}

\begin{example}[Conscientious objection] \label{conscient} An assistant bot is asked to prepare and mail phishing emails to a database of users. It refuses to do so because it is the wrong thing to do (immoral).  
 \end{example}

\begin{example}[Clash of autonomy]\label{autonomy}  An assistant robot is asked to get the home owner a beer. The robot refuses to comply because the human is an alcoholic. The example is from \cite{9812265}. 
\end{example}

\begin{example}[Not my job]\label{job} A cleaner robot is asked to clean a spill on the floor. It refuses to do it because according to the schedule, another cleaner robot has that responsibility and is scheduled to clean that floor later. The example is from \cite{Fisher25:recklessness}.   
\end{example}

\section{Example analysis}
Let us assume that we want some quasi-Asimov-rules-of-robotics priority on intelligent machine behaviour desiderata: first their operation must be safe, second it should further human set goals and third it should be efficient. 

We can consider the grounds on which the non-compliance is justifiable in each example. Examples~\ref{safety}-\ref{inable} have objective justification, namely there is evidence that non-compliance is necessary.  Examples~\ref{emergency}-\ref{autonomy} have a normative justification, there exists some norm  (lawfulness, ethics) that the machine is complying to that justifies the non-compliance with the command. In   Example ~\ref{job}, the non-compliance is practical. Namely it is more efficient for the machine's goals to not be derailed.  \citet{ijcai2023p33} argue that these three grounds -- objective (empirically true), subjective (based on beliefs and intentions), and practical (justified
in the social context) -- are due to \citet{Habermas1984}. 

\citet{FSS1511709} do not consider grounds for refusing a goal but a hierarchical chain of tests for adopting a goal: (1) Knowledge: does the robot know how to do X?; (2)  Capacity: is the robot physically able to do X now?;  (3) Goal priority and timing: is the robot able to do X right now?;  (4) Social role and obligation : is the robot obligated based on its
social role to do X?; (5) Normative permissibility : Does it violate any normative
principle to do X? 
 
The criteria (1) and (2) are safety or objective requirements, (3) is a practical requirement (in our rather than Habermas' sense of the word) whereas, (4) and (5) are normative requirements (legal and ethical). 

We can also observe that there are three categories of examples with respect to how able the user should be to accept the non-compliance. The first group are Examples~\ref{unavaialable}~and~\ref{inable} where the machine does not really have a choice because the task cannot be completed. In the second group are Examples~\ref{safety},~and~\ref{priority} where complying would lead to direct serious harm to the user or someone else. In this case the human operator should not have a choice to command the machine.  The Examples~\ref{emergency}-\ref{job},  we ordered in what can be seen as an ascending order of how acceptable it would be for a person to not agree with the refusal to comply for the machine.  

When \citet{coman2018ai} consider rebellious behaviour, they consider four stages of the process: pre-rebellion (the agent assesses changes in the environment and the behaviour of other agents),  rebellion deliberation (the agent assesses motivating, supporting, and inhibiting factors to trigger rebellion), rebellion execution (rebellion is triggered as a result of rebellion deliberation), and post-rebellion (observing the reactions to the act of rebellion).  

\begin{figure} 
    \centering
    \includegraphics[width=.9\linewidth]{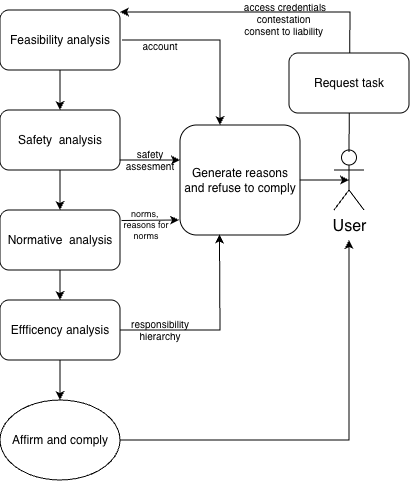}
    \caption{Request compliance life-cycle}
    \label{fig:diagram}
\end{figure} 

Because we are focused on non-compliance, the pre-rebellion phase for us is simply having agents designed in such a way that they consider a command when being issued one rather than directly execute it.  What follows is deliberation and decision-making following the different reasons that the machine can have to not comply.  Non-compliance execution is communicating acknowledgement of the command and either acceptance or refusal. The refusal is necessarily accompanied with possibly refutable grounds for refusal. The post non-compliance phase would consist of a possible override from the human in response to con-compliance which would trigger another non-compliance deliberation phase. 
From this very short analysis we can construct the following simplistic life-cycle of a refusable command, depicted  in Figure~\ref{fig:diagram}. 

The first step in considering whether to comply with a request is to evaluate whether the request is feasible. The agent may attempt to execute the task and fail, without being able to give reasons as to why the failure has happened. The next three steps however,  require an analysis which would produce reasons to refuse the user request.

\section{Understanding Safety}  
The safety paradox of autonomous systems is tensioned on whether machines should obey humans unconditionally, or whether disobedience itself can be a critical safety feature. A truly safe autonomous system must sometimes refuse instructions, whether those commands stem from human error, malicious intent or incomplete understanding of the situation. 
 
 We can consider at least three categories of unsafe operation: unsafe for the user, unsafe for the environment and unsafe for the machine. This safety triad creates competing priorities that complicate disobedience in autonomous systems. 
 
 Protecting the user seems like it should be the most important aspect but it creates dilemmas. Should the machine refuse a command that endangers the user even if the user accepts the risk? For example, a medical robot might refuse a risky surgical approach that a surgeon deems necessary. At what point does protecting users from themselves become counterproductive?
 We would argue that non-compliance due to safety concerns for the person who issues the command should be refutable. 

 In regard to environment safety, we can think of situations where, for example, an industrial robot must protect nearby workers, not just its operator \cite{hanna2022deliberative}. In some ways, environmental safety provides clearer grounds for non-compliance because the affected parties (nearby workers) cannot consent to, or override, the risk in the same way as an operator. This is also reason to not allow for a user to refute the non-complience and force execution. 

Self-preservation in machines raises philosophical questions. Should a system refuse commands that damage it? This has implications for the other two categories, a damaged machine cannot protect its user or the environment. For example, an industrial robot that operates beyond its specifications becomes unreliable, potentially endangering workers \cite{lee2021critical}. Machine self-protection maintains safety. However, systems must accept damage to prevent human harm --- the autonomous vehicle should sacrifice itself rather than hit a pedestrian \cite{evans2020ethical}. The difficulty lies in balancing how much machine preservation is justified when it conflicts with user convenience or minor environmental impacts. We would argue that refutation of the non-compliance should be possible in some cases. We do however, now run into the question of ownership: whose machine is this to damage? 

In essence, the machine needs  some kind of flexible safety hierarchy. Sometimes user safety dominates (medical emergencies), sometimes environmental safety does (preventing accidents) and sometimes machine capability matters the most (a rescue robot that is capable of reaching survivors in a collapsed structure). The context determines this priority so the machine must dynamically assess and weigh these priorities against each other.

 This balancing act raises the question: should humans retain the power to override the machine's safety calculus? The answer isn't obvious. Human overrides can introduce their own safety issues. For example, an operator might override to protect themselves while endangering the environment, including other workers. They might sacrifice machine integrity for immediate user benefit, degrading long-term safety capacity.

\section{Understanding Normative Obedience}
The fields of AI alignment \citep{gabriel_artificial_2020,ji_ai_2025} and machine ethics \citep{Anderson2007} are both concerned with engineering autonomous agents and AI systems to adhere to social, ethical and legal norms. Although both of these fields consider the compliance with norms (and values) as a goal of its own, clearly normative conflicts will arise. That is, the machine will be asked to do a task that violates a specified norm or erodes a stated value. Since much has already been written on this topic, we will shortly focus just on the question: What is the right way to consider a person coercing an autonomous machine to do the wrong thing? Assuming that we consider machines, even autonomous machines to, first and foremost, be tools that extend the will of persons then that person should be able to require a machine to violate norms. The necessary conditions to do so responsibly is to: understand that norm violation will occur and to assume liability for the violation. 

In the general process that \citet{FSS1511709} describe for robots to refuse commands, they do not consider the option that a robot may have their own schedule and plan with which the request from the user might be interfering. We consider this case in our life-cycle analysis. 

\section{Understanding Task Priority}
 The autonomous machine might choose to refuse a command \textbf{not} because it deems it unsafe or due to a normative conflict,  but because either (a) it has something it believes \emph{more important} to do, or (b) it believes that, while it \emph{could} act, another agent (human or system) will tackle this task within a reasonable time-frame. 

One way to capture this task priority is to represent hierarchies of responsibilities.
This includes both a hierarchy concerning how responsible each agent is for an activity/action and a hierarchy of how each agent views its responsibilities. Simply put, the agent is responsible for several actions/outcomes and may be responsible to a range of other agents. In deciding how to proceed, the agent weighs its responsibilities and can choose to \emph{not comply with} a user command if it deems the request to not be ``as important'' as at least one of its other responsibilities. Furthermore, even when an agent \emph{could} comply with a request, it can choose to \emph{not comply} if it  believes another agent is capable and is \textbf{more responsible}. And therefore the other agent will/should tackle this~\cite{Fisher25:recklessness}, as is illustrated in Example~\ref{job}. 

\section{Understanding Reasons}
Figure~\ref{fig:diagram} clearly illustrates that all non-compliance should be accompanied by the reasons that justify it. This information would allow the user to understand whether the machine is doing the right thing, whether an override is possible and also rational, as well as possible liability that the user would be assuming if choosing to force compliance.  In Table~\ref{tab:refute} we summarise the cases for allowing a user to override non-compliance.  
\begin{table}[h]
\centering
\caption{Overview of non-compliance reasons}\label{tab:refute}

\begin{tabular}{llc}
\toprule
\textbf{Reason} &  & \textbf{Refutable by User} \\
\midrule
Feasibility &   & no  \\
\midrule
\multirow{3}{*}{Safety} & Unsafe to user & yes \\
 & Unsafe to environment &  no \\
 & Unsafe for machine &  upon emergency  \\
\midrule
\multirow{3}{*}{Normative issues} & Unethical & yes \\
& Antisocial & yes \\
 & Illegal &  yes  \\
\midrule
Efficiency &   & upon emergency \\
\bottomrule
\end{tabular}
\end{table}

What we would need to design are two reasoning channels. In the first channel, the non-compliant agent provides sufficient justification for refusing to comply. Non-compliance due to feasibility requires an account  \cite{DBLP:conf/aiia/BaldoniBM020}, that is a trace of what the agent has attempted to do and what is observed as a result.   Justifications differ from explanations and accounts in that they need to demonstrate not only why a decision is made but also why another decision would not have been possible \cite{KASACHKOFF1988}. The justifications of refusals to comply on the grounds of safety need to be supported with a safety analysis, whereas those on the ground of efficiency need to be supported with sufficient details on the hierarchies of responsibilities. Normative non-compliance needs to be grounded in both the norms that stand to be violated, but also in the reasons for which those norms exist in the first place \cite{Bench-Capon2016}. 

In the second channel, the user needs to supply its access authorisation for override and consent to assume liability, particularly in the cases which we labelled emergency override in Table~\ref{tab:refute}. The override in the cases of safety and efficiency can be reduced to communicating that one has authority, and consents to assuming liability. However, in the case of normative non-compliance and override, there contestation of the user need to be structured in a form of a dialogue \cite{DignumMN0ST25,10.24963/kr.2024/83}, requiring further sufficient and necessary requirements to terminate the dialogue.   

\section{Engineering non-compliance}

Questioning every command given by the user may not be the most efficient operational mode for machines. The machine needs to be designed so that non-compliance is considered only when it is actually reasonable to do so. We sketch out three different ways in which the machine can be engineered to deliberate about non-compliance.

\textbf{Deliberately non-compliant}  are those designed to always refute a command if it comes from a specific pre-defined set. The set membership can be characterised by context, type of task or a   type of user. For example, the machine does not comply with requests issued while it is operating in an unsafe environment, tasks that require accessing private information, or tasks request from a person named Michael. This is a simple  approach that also allows for the justification and override conditions to be simplified. It is, however, not very robust.   

  \textbf{Predictably non-compliant} are those designed to follow a reasoning mechanism and/or a priority threshold of outcomes where, when breached, means the machine will disobey. This pipeline can follow a feasible $\rightarrow$ safe $\rightarrow$ normative $\rightarrow$ efficient pipeline of checks, not always but when some thresholds are reached. For example, when the command interrupts an ongoing task, when there is some aspect of the request that flags a normative concern, when the battery is low etc.  In addition to the examples above, we can also consider a dynamic evaluation of priorities. For example, in the driverless car scenario, if there have been too many accidents when human driver has taken control at night, then the machine would not comply with request to let them drive at night.
  
  \textbf{Learnt non-compliance}  can be achieved by initially randomly selecting some commands to consider through the $\rightarrow$ safe $\rightarrow$ normative $\rightarrow$ efficient pipeline and attempting to extract from it patterns of contexts, tasks and users for which non-compliance should be considered. In effect this means self-generating the set that is pre-defined in the Deliberately Non-compliant architecture. 

 \section{Conclusions}
 The ability to not-comply with a user request is a necessary skill for autonomous intelligent machines. Non-compliance is a special skill in the pantheon of rebellious agent behaviour and requires its own attention, which in contrast the norm obedience has received in the multi-agent systems community. The difference between compliance and obeying a norm is that norms are grounded in values or reasons and their violation is monitored and sanctioned by a designated mechanisms. This structures the reasoning about disobedience. A user does not have to justify its request and has limited options to use sanctions as incentives to ensure compliance. Safety and efficiency concerns require different justification than for example normative conflicts. 
We here examine the pathways for engineering responsibly non-compliant machines. We argue that responsible non-compliance requires that the machine knows when to consider non-compliance, how to justify the decision to not comply and when to allow for the user to override its non-compliance. We further argue that security and liability concerns need special attention in non-compliance. Future work, clearly is to unpack the sketch we here present into a responsibly non-compliant agent architecture. 


\begin{acks}
If you wish to include any acknowledgments in your paper (e.g., to 
people or funding agencies), please do so using the `\texttt{acks}' 
environment. Note that the text of your acknowledgments will be omitted
if you compile your document with the `\texttt{anonymous}' option.
\end{acks}



\bibliographystyle{ACM-Reference-Format} 
\bibliography{sample}


\end{document}